# A Multi-Level Causal Intervention Framework for Mechanistic Interpretability in Variational Autoencoders

Dip Roy[1*], Rajiv Misra[1], Sanjay Kumar Singh[2], Anisha Roy[3]

[1]*Department of Computer Science and Engineering, Indian Institute of Technology, Patna, India*

[2]*Department of Computer Science, Rajarshi School of Management Technology, Varanasi, India*

[3] *Department of Electronics and Communication Engineering, Jaypee Institute of Information Technology, Noida*

*Corresponding author: dip_25s21res37@iitp.ac.in

## Abstract

To establish a basis for understanding how generative models represent and transform data is an essential problem in the field of deep learning interpretation. Although, the mechanistic interpretation of discriminative architectures (e.g., transformers) has produced substantial new insights about these types of systems, there has been relatively little work on mechanistic interpretation of variational autoencoders (VAEs). VAEs are widely used for representation learning; this paper presents the first general-purpose multilevel causal intervention framework for the mechanistic interpretation of VAEs. The framework includes four basic manipulation types: input manipulation, latent-space perturbation, activation-patching, and causal-mediation-analysis. This paper also defines three new quantitative metrics that measure various characteristics of VAE internal representations that are not measured by existing disentanglement metrics alone: Causal-Effect-Strength (CES), intervention specificity, and circuit modularity. We conducted the largest empirical study to date of the causal mechanisms of six VAE architectures (standard VAE, β-VAE, FactorVAE, β-TC-VAE, DIP-VAE-II, and VQ-VAE) over five different benchmark datasets (dSprites, 3DShapes, MPI3D, CelebA, and SmallNORB) using three different random seeds per architecture and dataset pair, leading to a total of 90 independent training runs. Our empirical results reveal several new findings: (i) a consistent within-dataset negative correlation between CES and DCI disentanglement, which we refer to as the CES-DCI trade-off; (ii) that the KL reweighting mechanism of β-VAE can cause a capacity bottleneck for β-VAE when the number of generative factors in a model approaches its latent dimensionality, causing a degradation of disentanglement performance on complex datasets; (iii) that no single VAE architecture will be best for all of the five datasets tested — the optimal choice of VAE architecture is dependent upon the structure of the dataset being modeled; and (iv) that CES-based metrics applied to discrete latent spaces (VQ-VAE) yield near zero values, indicating a critical limitation of continuous-intervention methods for analyzing discrete latent spaces. These results provide both a theoretical foundation and empirical evaluations of the interpretation of generative models.

**Keywords:** Mechanistic interpretability; Variational autoencoders; Causal interventions; Disentanglement; Circuit analysis; Causal effect strength

# 1. Introduction

Deep learning has been very successful in a wide variety of areas but the processes that they use to make decisions are still somewhat mysterious. Mechanistic interpretability is an area of research that studies what mechanisms (how) deep learning models process and represent information through the different parts of the model and the paths that the computation takes to get from input to output [1],[2]

Substantial advances have been achieved in developing an understanding of discriminative models — especially those employing transformer architecture. Inductive head [2], indirect object identification circuitry [3], and a variety of feature visualization tools [4] were developed to provide insight into the way each of these model architectures accomplishes particular tasks. Generative models, which include a wide variety of autoencoder architectures — including VAEs [5] — are currently receiving very little attention within the mechanistic interpretation community, although they represent a crucial component for many applications of unsupervised learning and representation discovery.

VAEs learn to map high-dimensional observations into structured latent spaces from which the original data can be reconstructed. Several architectural variants have been proposed to encourage disentangled representations, where distinct latent dimensions correspond to independent generative factors. These include β-VAE [6], FactorVAE [7], β-TC-VAE [8], DIP-VAE [9], and VQ-VAE [10]. While existing disentanglement metrics such as the DCI framework [11], Mutual Information Gap (MIG) [8], and SAP [12] evaluate the quality of learned representations, they treat the model as a black box, measuring input–output relationships without revealing the internal mechanisms through which disentanglement emerges or fails.

This paper bridges the gap between disentanglement evaluation and mechanistic understanding by developing a causal intervention framework that probes VAE internals at multiple levels. Rather than simply measuring whether a VAE has learned disentangled representations, our framework reveals *how* and *why* different architectures achieve varying degrees of disentanglement, and what causal costs these mechanisms impose on representation strength.

Specifically, we address the following research questions:

**RQ1:** How does the Causal Effect Strength (CES) of latent dimensions relate to traditional disentanglement metrics across different VAE architectures, and is this relationship consistent across datasets?

**RQ2:** Does the effectiveness of disentanglement penalties (KL reweighting, adversarial TC, covariance regularization) depend on the relationship between the number of generative factors and latent dimensionality?

**RQ3:** Can causal intervention metrics reveal mechanistic differences between VAE architectures that traditional disentanglement metrics cannot distinguish?

**RQ4:** How do continuous causal interventions behave when applied to discrete latent representations (VQ-VAE), and what does this reveal about the limitations of intervention-based interpretability?

Our contributions are as follows. First, we develop a multi-level causal intervention framework comprising input interventions, latent space perturbations, activation patching, and causal mediation analysis tailored for VAE architectures. Second, we introduce quantitative metrics—CES, intervention specificity, and circuit modularity—that capture complementary aspects of VAE representations beyond what existing disentanglement metrics measure. Third, we present the most comprehensive empirical study of VAE mechanistic interpretability to date, evaluating six architectures across five datasets with 90 total training runs. Fourth, we identify the CES–DCI tradeoff, the β-VAE capacity bottleneck, and the failure of continuous intervention metrics for discrete latent spaces, each representing novel findings with practical implications for VAE design and evaluation.

## 2. Related Work

### 2.1 Mechanistic Interpretability

Mechanistic interpretability is concerned with reversing the process of computation made by neural networks. A fundamental paper by Olah et al. [1] illustrated how individual neurons and collections of neurons within convolutional networks create interpretable features that operate together in circuitry. Elhage et al. [2] established a mathematical framework for studying attention-based circuits within transformer networks; subsequently, researchers have found evidence for various mechanisms including indirect object recognition in GPT-2 [3], and "grokking" [13], which can be studied at the level of circuit analysis. Network Dissection [4] provides quantitative techniques for evaluating the extent to which individual neurons correspond to specific semantic objects. However, virtually all of the research that has been conducted using these techniques are focused on discriminative architectures. Generative architectures pose different challenges because their output is a high-dimensional reconstruction rather than a discrete prediction, and they have additional layers of representational complexity due to latent space.

Recent research is expanding upon a mechanistic perspective of Generative Models. Bau et al. [29] applied Network Dissection to GANs; they found neurons that controlled particular visual characteristics. Conwell et al. [30] analysed the organization of visual characteristics of VAE's by layer but were limited to only observing responses to input data as opposed to establishing causal relationships through intervention(s). The primary distinction between our work and their work lies in the fact that we have adopted a mechanistic approach based on intervention (as opposed to observation) for the study of internal workings of VAEs.

### 2.2 Disentangled Representation Learning

Disentanglement is the goal of many VAE versions - where every latent dimension represents one single variable and therefore independent from other dimensions. Higgins et al. [6] were the first to introduce the Beta-VAE - they added a penalty to the KL term (the part of the loss function that measures how close the learned distribution is to the standard normal) to encourage the learned distribution to be a product of independent factors. Kim and Mnih [7] then extended this by introducing Factor-VAE - using a discriminator to add another penalty to the loss function for the total correlation of the distribution of all the latent dimensions. In their version of the Beta-VAE, Chen et al. [8] broke down the KL term into three components: total correlation, mutual information, and dimension-wise KL. This provided a formal way of understanding which part of the KL term contributes most to the disentanglement. Kumar et al. [9] developed DIP-VAE - adding a new penalty to the loss function based on the difference between the moments of the aggregated posterior and a factorised prior; the penalty is based on a covariance regularisation term. Van den Oord et al. [10] used VQ-VAE - replacing continuous latent variables with discrete codebook entries. This was a very different approach to structure representation learning.

Locatello et al., [14], in their extensive evaluation study demonstrated that disentanglement methods are highly sensitive to both hyperparameter selection as well as random seed assignments, and therefore, no single method outperforms all other methods across all environments regardless of architecture type. The authors demonstrated the necessity of adopting multi-seed evaluations (a methodology that we employ), as well as the need for developing a deeper mechanistic understanding regarding why certain disentanglement methods perform better than others. Zhao et al., [27] proposed InfoVAE which balances reconstruction performance with information-theoretic regularized loss functions. Do and Tran [28], analyzed theoretically, the evaluation of disentanglement methods using multiple metrics, thereby illustrating that these multiple metrics measure different representations within the data space. Our work provides complementary mechanistic explanations (using causal intervention) to explain the previously identified architectural and data dependent relationships that were previously documented, but unexplained, by prior studies.

## 2.3 Causal Methods in Neural Network Analysis

A causal approach to analyzing internal workings of neural networks has been successful using Pearl's causal inference [26]. It was shown that correlating observed data from neural network components is not sufficient to understand how they work mechanistically, but rather by intervening on each component (i.e., by manipulating one or more components), we will be able to identify causal relationships. Geiger et al. [15] have provided a method for developing causal abstractions based on an interpreter which maps model computation to causal models and thus provides a mathematical link between causal inference theory and neural network analysis. Meng et al. [16] demonstrated that through the use of causal tracing methods that factual associations contained within large language models can be located at specific layers and neurons.

Vig et al. [17] used causal mediation analysis to study gender bias in NLP models, quantifying the extent to which specific attention heads mediate biased predictions. Activation patching, where activations from one input replace those from another at specific network locations, has been particularly effective for identifying causal pathways in transformers [3], [18]. In the generative model domain, Yang et al. [31] proposed CausalVAE, which incorporates causal structure into the generative process itself, while Suter et al. [32] studied robustly disentangled causal mechanisms focusing on invariance properties. Our work differs from CausalVAE in that we do not modify the generative architecture but instead develop post-hoc intervention tools that can be applied to any trained VAE. We differ from Suter et al. by focusing on internal circuit-level analysis rather than robustness properties of learned representations.

### 2.4 Positioning of This Work

Our framework occupies a unique position at the intersection of mechanistic interpretability, causal inference, and disentanglement evaluation. While existing disentanglement metrics [11], [8], [12], [28] evaluate representations through input-output statistics, and mechanistic interpretability tools [1], [2], [3] have focused primarily on discriminative models, our approach applies causal intervention techniques specifically designed for the VAE encoder-decoder architecture to provide both quantitative metrics (CES, specificity, modularity) and mechanistic explanations for how different architectural choices shape internal representations. This paper substantially extends our preliminary work, which examined three architectures on two datasets, to a comprehensive study of six architectures across five benchmarks with rigorous multi-seed evaluation.

## 3. Methodology

### 3.1 Problem Formulation

The goal is to develop a general and methodical way of identifying and evaluating how the causal paths of knowledge regarding generative factors are transmitted from the inputs into the encoded bottleneck in the latent space to reconstructed data as a result of training a VAE with an encoder $E\phi$ and decoder $D\theta$ on data generated by K independent factors of variation. In addition, we wish to identify which latent variables represent or relate to which generative factor(s), how much influence is exerted upon reconstruction by each latent variable, and how various architectural options influence these causal pathways.

### 3.2 Multi-Level Causal Intervention Framework

Our framework performs interventions at four levels of the VAE, each providing distinct insights into the model's internal mechanisms.

#### *3.2.1 Level 1: Input Interventions*

Input interventions systematically modify specific factors of the input data and track how these changes propagate through the network. For an observation x and a target factor f, we construct a modified input:

$$\tilde{x} = f_{intervene}(x, f, v) \quad (1)$$

that alters factor f to value v while keeping all other factors constant. We then measure the activation difference across the encoder:

$$\Delta z = E_\phi(\tilde{x}) - E_\phi(x) \quad (2)$$

to identify which latent dimensions and intermediate activations respond to changes in each factor. This level establishes the basic mapping between generative factors and network components.

### *3.2.2 Level 2: Latent Space Interventions*

Latent space interventions directly perturb individual dimensions of the encoded representation $z = E\phi(x)$. For each dimension i, we construct an intervened representation:

$$\tilde{z}_i = z \text{ with dimension } i \text{ set to value } v \quad (3)$$

by sweeping dimension i across the range [−3, +3] while holding all other dimensions fixed. We then measure the reconstruction difference:

$$\Delta x_{recon} = D_\theta(\tilde{z}_i) - D_\theta(z) \quad (4)$$

This level quantifies the causal influence of each latent dimension on the output and forms the basis for computing CES and specificity.

### *3.2.3 Level 3: Activation Patching*

Activation patching replaces the internal activations produced by one input with those from another at a specific layer l. For inputs $x_1$ and $x_2$, we define patched activations where a target neuron n is replaced:

$$\tilde{A}_l = A_l(x_1) \text{ with neuron } n \text{ set to } A_l(x_2)[n] \quad (5)$$

The resulting modified reconstruction reveals which neurons at each layer are causally responsible for processing specific aspects of the input. This level identifies the computational pathways, or circuits, through which information flows.

### *3.2.4 Level 4: Causal Mediation Analysis*

Causal mediation analysis quantifies the extent to which specific components mediate the total effect of an input intervention. For a factor change from x to x̃, we compute the total effect $TE = f(x) - f(\tilde{x})$ and the mediated effect ME_C through component C by replacing only C's activations from the intervened input while keeping all other activations from the original. The mediation strength ME_C/TE reveals which layers and channels serve as the primary information conduits for each factor.

## 3.3 Proposed Metrics

### *3.3.1 Causal Effect Strength (CES)*

CES measures the magnitude of change in the decoder output caused by a unit perturbation of a single latent dimension. For latent dimension i:

$$CES(i) = E_x[\|D_\theta(z) - D_\theta(\tilde{z}_i)\|^2] \quad (6)$$

where z̃_i represents the latent vector with dimension i set to a fixed intervention value. Higher CES indicates that the dimension exerts a stronger causal influence on the reconstruction. The average CES across dimensions characterizes the overall representational strength of the model.

### *3.3.2 Intervention Specificity*

Intervention specificity quantifies how spatially localized the effects of a latent perturbation are:

$$S(i) = 1 / (H(p_i) + \varepsilon) \quad (7)$$

where p_i is the normalized distribution of pixel-wise squared differences across the output and H(p_i) is the Shannon entropy of this distribution. Higher specificity indicates that the perturbation affects a focused region rather than causing diffuse global changes, suggesting more semantically meaningful encoding.

### *3.3.3 Circuit Modularity*

Circuit modularity measures the degree to which different computational circuits in the network are specialized for processing distinct factors. For a layer with N neurons and K interventions:

$$M = 1 - (2 / K(K-1)) \Sigma_{i<j} |\rho(\Delta a_i, \Delta a_j)| \quad (8)$$

where Δa_i is the vector of activation changes for intervention i across all neurons and ρ is the Pearson correlation coefficient. Lower average correlation (higher M) indicates that different interventions activate largely non-overlapping sets of neurons, reflecting greater functional specialization.

## 3.4 Relationship to Existing Metrics

Our proposed metrics build upon and extend several existing approaches in the interpretability and disentanglement literature. Table 1 situates CES, intervention specificity, and circuit modularity relative to established metrics, clarifying what each uniquely contributes.

**Table 1: Comparison of proposed metrics with existing approaches**

| Metric | Related Prior Work | Key Difference |
|---|---|---|
| CES (ours) | Latent traversal sensitivity [6]; Jacobian norm [33] | CES measures output-space L2 effect per dimension under fixed intervention range, not gradient-based; aggregated over samples, enabling cross-architecture comparison |
| Specificity (ours) | Spatial concentration in GAN dissection [4]; Receptive field analysis [29] | Operates on decoder output space rather than intermediate features; inversely related to entropy of pixel-wise effects, capturing semantic localization |
| Modularity (ours) | Representation similarity [34]; Network modularity [1] | Correlation-based but computed over factor-specific intervention responses rather than stimulus-driven activations; measures functional specialization for generative factors |

CES has two important differences with respect to Jacobian-based sensitivity measures [33]. The first difference is that CES utilizes finite difference based interventions on the interval [-3,+3] whereas infinitesimal gradients are utilized by Jacobian-based sensitivity measures. Therefore, CES represents a more accurate representation of how a model behaves under realistic perturbation magnitudes. The second difference is that CES is calculated as the expectation over the data distribution instead of at one operating point; thus, CES represents a more robust characterization of the causal role of each dimension. With regard to the traversal variance implicit within the prior disentanglement literature [6],[7], CES represents a standardized and dimension-specific measure of variance which can be used for direct comparisons between architectures and datasets.

Spatial concentration is another factor that can be used to support the specific type of interventions (as opposed to the metrics based on spatial concentration for GAN dissection [4],[29]) , because it is applied to the output of the decoder . Network Dissection is about measuring the degree of correspondence between forward pass statistics of neurons and their corresponding semantic concepts. Our Specificity Metric is also about measuring the spatial concentration of the causal effect of latent perturbations. Therefore, we provide a double perspective: dissection is asking "What does this neuron capture?" , while specificity is asking "How focused is this dimension's effect?".

Circuit modularity is an extension of representation similarity analysis [34], which can be applied from observational to interventional settings. Instead of generating a correlation matrix based on activity caused by stimuli; the circuit modularity measure generates a correlation matrix based on responses caused by interventions that are specific to each generative factor. This provides evidence for causal relations between network components and semantic factors as opposed to correlational relations.

### 3.5 Theoretical Basis for the CES–DCI Tradeoff

The empirically observed inverse relationship between CES and DCI within a dataset can be grounded in information-theoretic principles. Consider a VAE whose ELBO objective includes a KL regularization term weighted by β. The total information transmitted through the latent bottleneck is bounded by the aggregate KL divergence:

$$I(x; z) \leq \Sigma_i KL(q(z_i|x) \,||\, p(z_i)) \quad (9)$$

The optimization pressure reduces the channel bandwidth when β > 1. In response to this compression, the model suppresses those dimensions having lower MI to the data (moving their posterior toward the prior thereby producing near zero KL and near zero CES) and concentrates MI in the remaining "active" dimensions. The suppression of each remaining dimension's MI for a single factor is exactly the mechanism through which disentanglement occurs; i.e., it results in factorial representations at the expense of decreased total CES.

This is another example of how an argument can be connected with the information bottleneck principle [25], [35] because if we have a representation that best compresses the input and still has the most important information about a certain target variable then it will have a trade-off in terms of how well the data is compressed (the amount of low mutual information or low CES) and the quality of the predictions (how informative the representation is about the factors or how much DCI). The β parameter determines where on this trade-off curve the representation lies; therefore, if we want to favor better compression (more low CES), we would increase β so that there are less active dimensions for each factor (and thus more low CES) and the alignment of each dimension to its factor is stronger (thus more high DCI).

Architectures that don't suppress dimensionality — such as FactorVAE (penalizing the total correlation in the loss function but allowing each of the KL terms to be un-constrained) or DIP-VAE (penalizing a mismatch in the covariance) — show either a much reduced or completely absent CES–DCI trade-off. This is because both architectures allow a reorganization of information from one dimension to others, which doesn't reduce the overall amount of information being passed through the model. It is this ability to reorganize information across dimensions that allows FactorVAE to maintain a high degree of CES and at the same time achieve moderate DCI improvements. Empirical evidence also supports the theoretical prediction that the trade-off will depend upon the mechanism used for regularization. The within-dataset correlation between CES and DCI was $r = -0.95$ for dSprites where β-VAE dominated the variation, however it was non-significant for CelebA where there were no significant improvements in disentanglement to measure.

# 4. Experimental Setup

## 4.1 Architectures

We evaluate six VAE architectures, each employing a distinct mechanism for structuring the latent space. All models share a common convolutional encoder-decoder backbone detailed in Table 2. The architectural variants and their regularization-specific hyperparameters are: (i) Standard VAE [5] with no additional regularization; (ii) β-VAE [6] with β = 4.0; (iii) FactorVAE [7] with adversarial TC penalty $\gamma = 10.0$ and a 2-epoch discriminator warmup; (iv) β-TC-VAE [8] with equal decomposition weights ($\alpha = \beta = \gamma = 1$); (v) DIP-VAE-II [9] with covariance penalty ($\lambda_{od} = 10$, $\lambda_d = 100$); and (vi) VQ-VAE [10] with 512 codebook entries and commitment cost 0.25.

**Table 2: Shared encoder-decoder architecture (all continuous VAE variants)**

| Component | Layer | Channels | Kernel | Stride | Output |
|---|---|---|---|---|---|
| Encoder | Conv2D + ReLU | 32 | 4×4 | 2 | 32×32 |
| Encoder | Conv2D + ReLU | 64 | 4×4 | 2 | 16×16 |
| Encoder | Conv2D + ReLU | 128 | 4×4 | 2 | 8×8 |
| Encoder | Flatten + Linear | — | — | — | $\mu, \log\sigma^2 \in \mathbb{R}^{10}$ |
| Decoder | Linear + Reshape | 128 | — | — | 8×8 |
| Decoder | ConvT2D + ReLU | 64 | 4×4 | 2 | 16×16 |
| Decoder | ConvT2D + ReLU | 32 | 4×4 | 2 | 32×32 |
| Decoder | ConvT2D + Sigmoid | C_in | 4×4 | 2 | 64×64 |

*C_in = 1 for grayscale datasets (dSprites, SmallNORB), 3 for RGB (3DShapes, MPI3D, CelebA). Total parameters: ~500K for continuous variants, ~2.1M for FactorVAE (includes discriminator). VQ-VAE replaces the μ/logσ² projection with vector quantization.*

We acknowledge that the hyperparameter values ($\beta = 4$, $\gamma = 10$) were not individually tuned per dataset. This choice was deliberate: our objective is to compare the mechanistic effects of different regularization mechanisms under standardized conditions, not to achieve maximum disentanglement for each variant. The hyperparameter sweep in Section 5.5 explores sensitivity to these choices on dSprites, demonstrating that our findings are robust across a range of settings. Per-dataset tuning would optimize absolute DCI values but would confound the mechanistic comparison by introducing dataset-specific biases into each architecture's behavior.

## 4.2 Datasets

We conduct experiments across five established benchmarks that span a range of complexities and factor structures.

**Table 3: Dataset characteristics**

| Dataset | Samples | Factors | Channels | Factor Types |
|---|---|---|---|---|
| dSprites [19] | 737,280 | 5 | 1 (binary) | shape, scale, orientation, posX, posY |
| 3DShapes [20] | 480,000 | 6 | 3 (RGB) | floor hue, wall hue, object hue, scale, shape, orientation |
| MPI3D [21] | 1,036,800 | 7 | 3 (RGB) | color, shape, size, height, background, horiz., vert. |
| CelebA [22] | 162,770 | 10 (binary) | 3 (RGB) | binary facial attributes (correlated) |

| SmallNORB [23] | 24,300 | 5 | 1 (gray) | category, instance, elevation, azimuth, lighting |
|---|---|---|---|---|

These datasets were chosen to represent increasing complexity: dSprites provides a controlled setting with binary images and independent factors; 3DShapes introduces RGB rendering with well-separated factors; MPI3D contains real-world-like images with more factors than latent dimensions; CelebA presents correlated real-world attributes; and SmallNORB provides a small-scale, grayscale object recognition benchmark.

## 4.3 Training Protocol

All models are trained for 30 epochs using the Adam optimizer with a learning rate of $10^{-3}$, weight decay of $10^{-5}$, and cosine annealing scheduling. We use a batch size of 512 on an NVIDIA L40S GPU (48 GB VRAM). Each model–dataset combination is trained with 3 random seeds (0, 1, 2), yielding 90 independent training runs. Results are reported as mean ± standard deviation across seeds. After training, we compute five standard disentanglement metrics (DCI [11], MIG [8], SAP [12], β-metric [6], FactorVAE metric [7]) and our proposed causal intervention metrics. Experiment tracking is performed with Weights & Biases [24].

The overall computational cost of this experiment was about 22 GPU hours to train and compute the metrics and perform the causal intervention. The training time was about 18 GPU hours and the other time was about 4 GPU hours as shown in figure 11; Training times were consistent across all architectures and we confirmed convergence after 30 epochs by looking at validation loss for all architectures and dataset combinations, and also we saw a stabilization of the relative validation loss (< 1% for the last 5 epochs). These training conditions are consistent with those of the larger scale evaluations of Locatello et al. [14] and Mathieu et al. [33] who showed that disentanglement metrics converge rapidly before 50 epochs for these models and datasets.

## 4.4 Reproducibility

To support reproducibility, we report: (i) the complete encoder-decoder architecture (Table 2); (ii) all regularization hyperparameters (Section 4.1); (iii) the training protocol including optimizer, learning rate, and scheduling; (iv) all random seeds (0, 1, 2); and (v) exact dataset preprocessing steps. All experiments are tracked with Weights & Biases [24] and the full experimental results are provided as JSON artifacts. Code is available at https://github.com/diproy123/MI_VAE

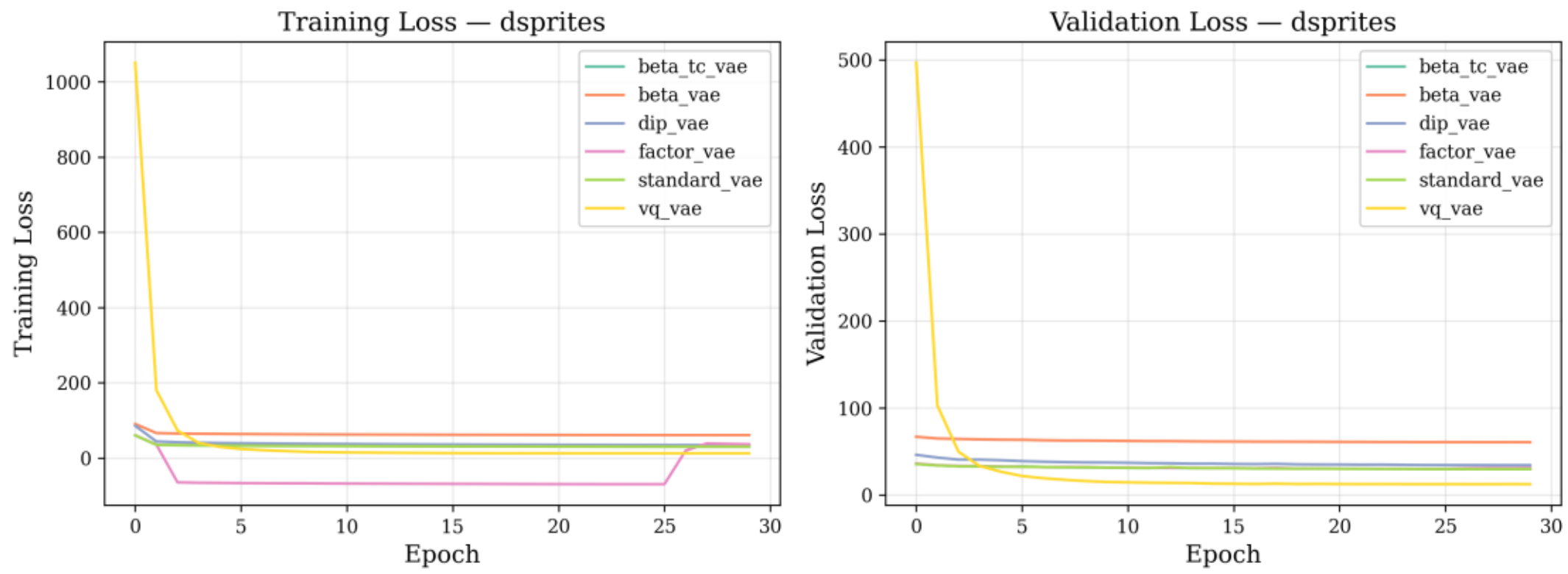

*Fig. 11. Training loss curves on dSprites for all six architectures across 30 epochs. All continuous VAE variants converge by epoch 20–25. FactorVAE shows negative training loss due to the adversarial TC penalty term (validation loss remains positive and converges normally). VQ-VAE achieves the lowest reconstruction loss due to its discrete codebook capacity.*

# 5. Results and Discussion

## 5.1 Cross-Architecture Comparison (RQ1, RQ3)

Table 4 presents the primary results across all five datasets and six architectures. We report DCI disentanglement and average CES as the principal metrics, with MIG included for cross-reference. Figure 1 contrasts the latent traversals of β-VAE and Standard VAE on 3DShapes, visually illustrating the difference between disentangled and entangled representations.

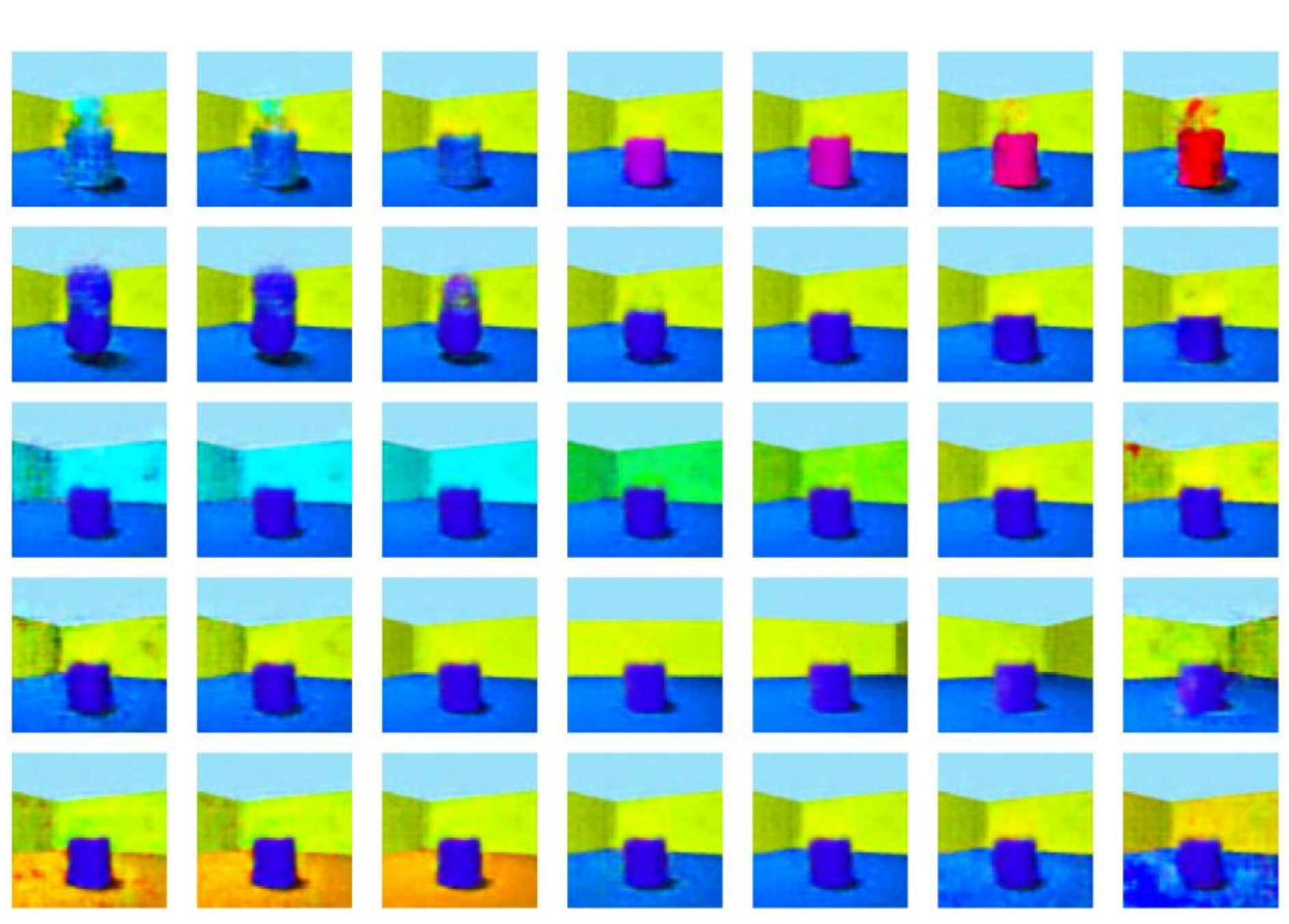

*Fig. 1. Latent traversals for β-VAE on 3DShapes. Each row corresponds to a single latent dimension swept from −3 to +3. Individual rows control distinct factors (object hue, wall hue, floor hue, scale, orientation), demonstrating disentangled encoding (DCI = 0.805).*

Latent Traversals — standard_vae on shapes3d

*Fig. 2. Latent traversals for Standard VAE on 3DShapes. Multiple factors change simultaneously within single rows, indicating entangled representations (DCI = 0.669). Compare with Fig. 1 to observe the visual difference in disentanglement quality.*

**Table 4: Comprehensive results across datasets and architectures (mean ± std over 3 seeds)**

**Panel A: dSprites**

| Model | DCI ↑ | MIG ↑ | SAP ↑ | CES | Modularity (μ) |
|---|---|---|---|---|---|
| Standard VAE | 0.100 ± 0.019 | 0.005 ± 0.003 | 0.018 ± 0.009 | 8.231 | 0.132 |
| β-VAE (β=4) | 0.305 ± 0.129 | 0.118 ± 0.078 | 0.222 ± 0.110 | 5.024 | 0.001 |
| FactorVAE (γ=10) | 0.104 ± 0.019 | 0.015 ± 0.009 | 0.071 ± 0.022 | 7.688 | 0.032 |
| β-TC-VAE | 0.090 ± 0.016 | 0.008 ± 0.002 | 0.026 ± 0.007 | 8.234 | 0.090 |
| DIP-VAE-II | 0.171 ± 0.102 | 0.022 ± 0.023 | 0.088 ± 0.085 | 7.881 | 0.041 |
| VQ-VAE | 0.075 ± 0.008 | 0.003 ± 0.001 | 0.005 ± 0.002 | 0.166 | 0.009 |

**Panel B: 3DShapes**

| Model | DCI ↑ | MIG ↑ | SAP ↑ | CES | Modularity (μ) |
|---|---|---|---|---|---|
| Standard VAE | 0.669 ± 0.088 | 0.035 ± 0.025 | 0.069 ± 0.044 | 22.848 | 0.464 |
| β-VAE (β=4) | 0.805 ± 0.011 | 0.169 ± 0.007 | 0.224 ± 0.005 | 18.836 | 0.220 |
| FactorVAE (γ=10) | 0.749 ± 0.031 | 0.071 ± 0.046 | 0.091 ± 0.049 | 21.338 | 0.160 |
| β-TC-VAE | 0.681 ± 0.051 | 0.049 ± 0.036 | 0.083 ± 0.064 | 22.879 | 0.392 |
| DIP-VAE-II* | 0.867 ± 0.053 | 0.089 ± 0.050 | 0.231 ± 0.076 | 19.514 | 0.802 |
| VQ-VAE | 0.424 ± 0.012 | 0.001 ± 0.000 | 0.004 ± 0.002 | 9.424 | 0.005 |

**DIP-VAE-II: DCI computed from seeds 1–2 only (seed 0 produced numerical −∞ due to dead latent dimensions).*

**Panel C: MPI3D**

| Model | MIG ↑ | SAP ↑ | CES | Modularity (μ) |
|---|---|---|---|---|
| Standard VAE | 0.085 ± 0.008 | 0.131 ± 0.007 | 4.718 | 0.023 |
| β-VAE (β=4) | 0.049 ± 0.005 | 0.076 ± 0.003 | 2.615 | 0.003 |
| FactorVAE (γ=10) | 0.079 ± 0.008 | 0.110 ± 0.019 | 4.664 | 0.034 |
| β-TC-VAE | 0.088 ± 0.006 | 0.137 ± 0.005 | 4.803 | 0.028 |
| DIP-VAE-II | 0.105 ± 0.040 | 0.113 ± 0.036 | 4.089 | 0.009 |
| VQ-VAE | 0.018 ± 0.009 | 0.007 ± 0.004 | 1.678 | 0.001 |

*Note: DCI on MPI3D produced ∞/−∞ for multiple model–seed combinations due to dead latent dimensions. MIG and SAP are reported as primary metrics for this dataset.*

**Panel D: CelebA**

| Model | DCI ↑ | MIG ↑ | CES | Modularity (μ) |
|---|---|---|---|---|
| Standard VAE | 0.099 ± 0.006 | 0.004 ± 0.001 | 15.863 | 0.241 |
| β-VAE (β=4) | 0.087 ± 0.013 | 0.007 ± 0.002 | 14.853 | 0.352 |
| FactorVAE (γ=10) | 0.078 ± 0.008 | 0.004 ± 0.001 | 14.534 | 0.086 |
| β-TC-VAE | 0.081 ± 0.011 | 0.003 ± 0.001 | 15.738 | 0.167 |
| DIP-VAE-II | 0.073 ± 0.011 | 0.003 ± 0.002 | 15.868 | 0.676 |
| VQ-VAE | 0.052 ± 0.006 | 0.001 ± 0.001 | 2.274 | 0.001 |

**Panel E: SmallNORB**

| Model | DCI ↑ | MIG ↑ | CES | Modularity (μ) |
|---|---|---|---|---|
| Standard VAE | 0.268 ± 0.005 | 0.097 ± 0.023 | 3.399 | 0.008 |
| β-VAE (β=4) | 0.223 ± 0.005 | 0.115 ± 0.004 | 2.369 | 0.001 |
| FactorVAE (γ=10) | 0.243 ± 0.023 | 0.099 ± 0.027 | 3.186 | 0.007 |
| β-TC-VAE | 0.268 ± 0.007 | 0.124 ± 0.020 | 3.304 | 0.007 |
| DIP-VAE-II | 0.227 ± 0.018 | 0.099 ± 0.016 | 3.536 | 0.026 |
| VQ-VAE | 0.055 ± 0.003 | 0.002 ± 0.001 | 1.819 | 0.000 |

## 5.2 The CES–DCI Tradeoff (RQ1)

The CES–DCI tradeoff predicted in Section 3.5 is borne out in the experimental data. Figure 3 shows the per-dimension CES and specificity across all six architectures on dSprites, revealing distinct CES profiles. Table 5 quantifies the tradeoff: Standard VAE versus β-VAE across all datasets.

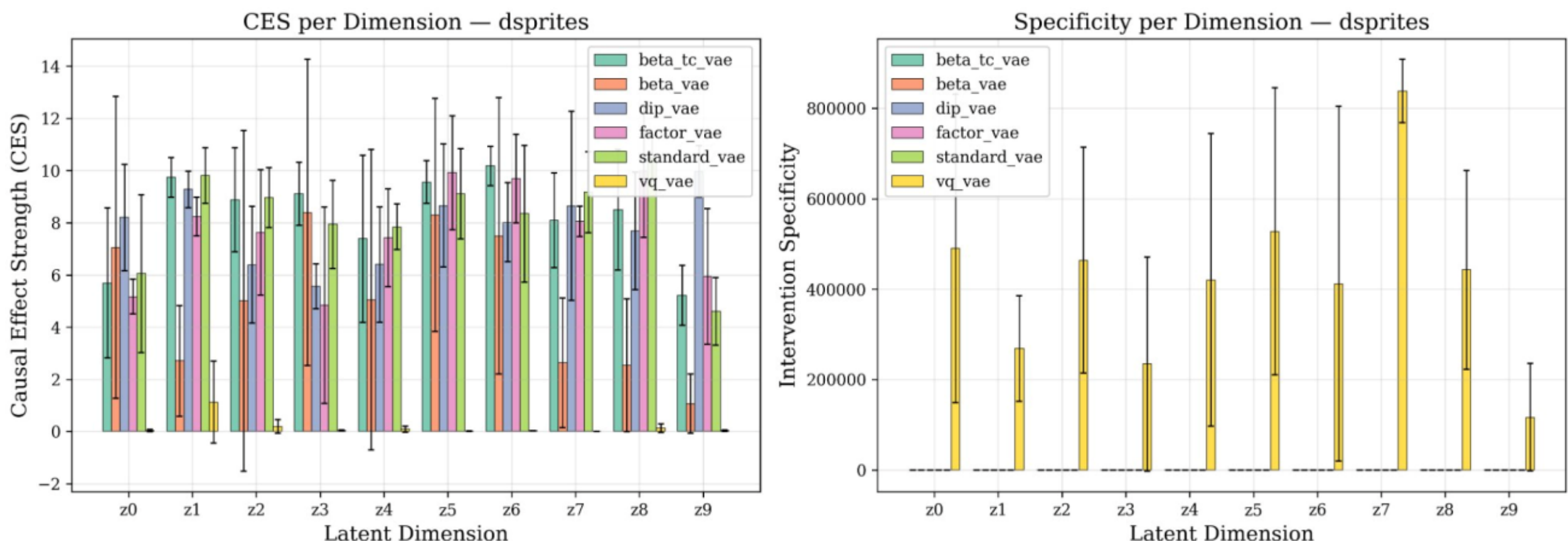

*Fig. 3. Per-dimension Causal Effect Strength (left) and Intervention Specificity (right) on dSprites. β-VAE (orange) shows reduced CES across most dimensions compared to Standard VAE (green), FactorVAE (pink), and β-TC-VAE (teal). VQ-VAE (yellow) shows near-zero CES but anomalously high specificity due to the discrete codebook.*

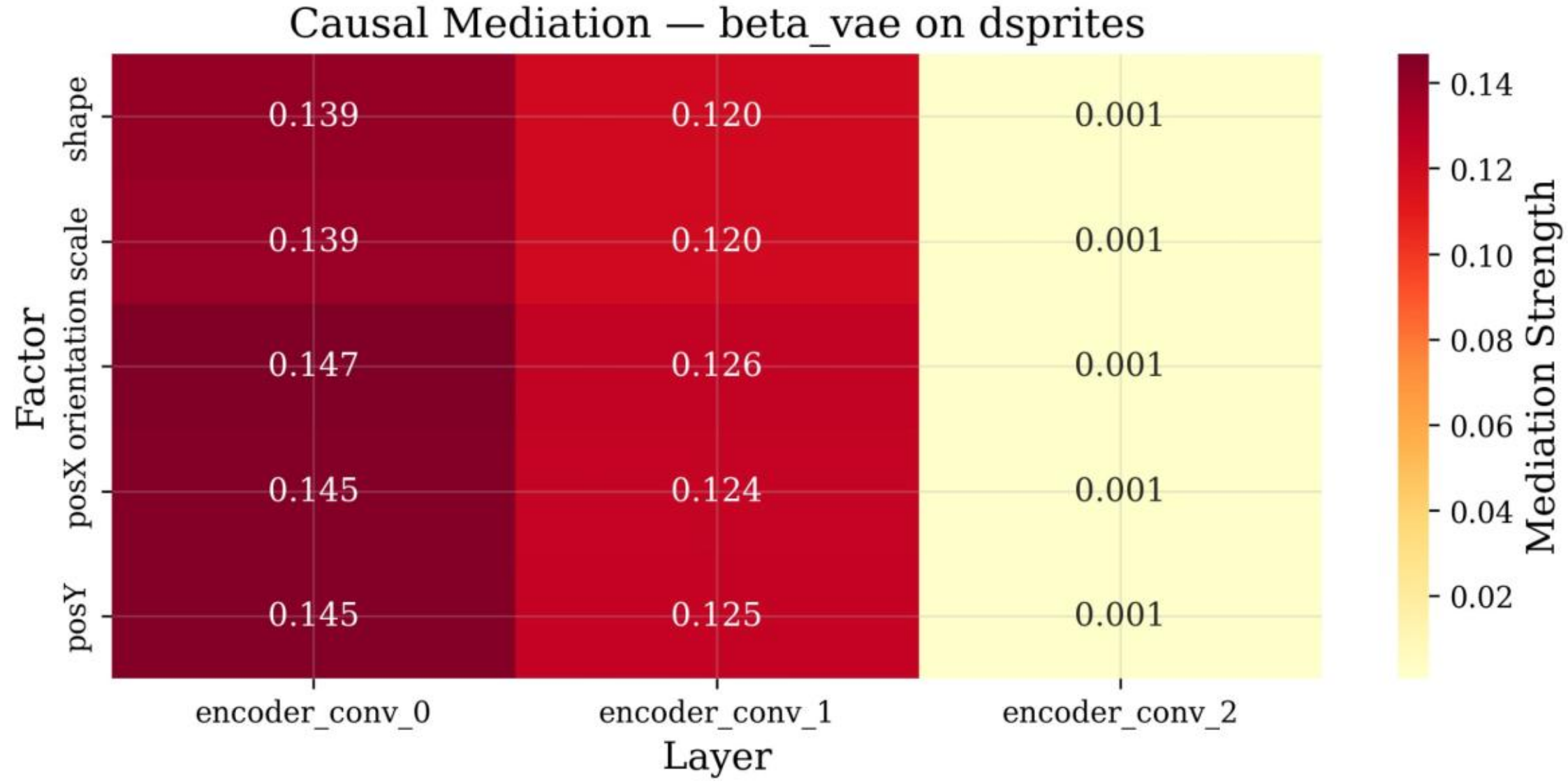

*Fig. 4. Causal mediation heatmap for β-VAE on dSprites. Rows represent generative factors; columns represent encoder layers. Mediation strength is concentrated in encoder_conv_0 and encoder_conv_1 (values ~0.12–0.15), dropping sharply at encoder_conv_2 (~0.001). This indicates that factor-specific information is primarily processed in early convolutional layers.*

**Table 5: CES–DCI tradeoff across datasets (β-VAE vs. Standard VAE)**

| Dataset | Factors | ΔDCI | ΔCES | DCI per CES unit |
|---|---|---|---|---|
| dSprites | 5 | +0.205 | −3.207 | 0.064 |
| 3DShapes | 6 | +0.136 | −4.012 | 0.034 |
| MPI3D | 7 | −0.151* | −2.103 | — |
| CelebA | 10 | −0.012 | −1.010 | — |
| SmallNORB | 5 | −0.045 | −1.030 | — |

**Negative ΔDCI indicates β-VAE performed worse than Standard VAE.*

On both dSprites and 3DShapes—datasets on which β-VAE was successful in improving disentanglement—we find that each unit of CES sacrificed results in an increase in DCI of 0.034–0.064 units. The process is interpretable as well: β-VAE's higher KL penalty will eliminate some latent dimensions altogether (thereby reducing their contribution to CES) and force the remaining dimensions (which remain active) to specialize for different individual factors (therefore increasing DCI). In this manner, we see an instance of "representational compression" — where the model sacrifices its capacity to express a wide variety of concepts for an increased ability to organize concepts structurally.

To rigorously test the CES–DCI relationship, we compute Pearson correlation coefficients within each dataset across five continuous VAE variants (excluding VQ-VAE). Figure 8 presents the scatter plot of CES versus DCI across all datasets (panel a) and the per-dataset correlation coefficients (panel b).

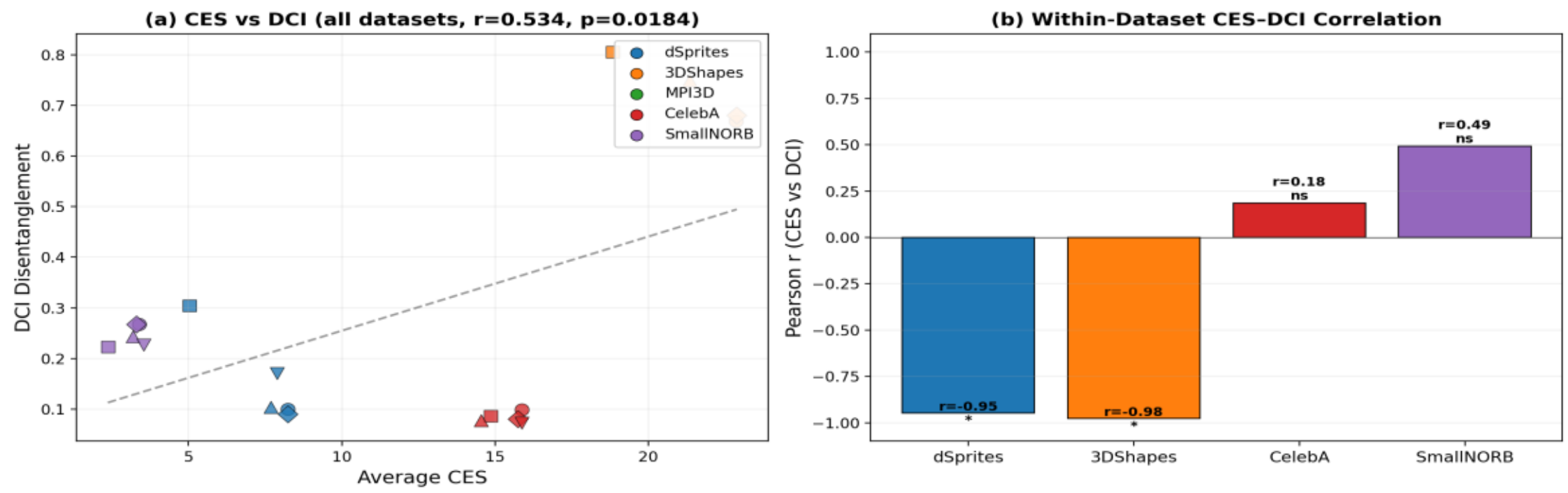

*Fig. 8. CES–DCI relationship analysis. (a) Scatter plot of average CES vs. DCI disentanglement across all datasets and continuous VAE architectures, showing a positive cross-dataset trend (r = 0.534, p = 0.018) driven by dataset complexity differences. (b) Within-dataset Pearson correlations reveal strong negative correlations on dSprites (r = −0.95, p = 0.014) and 3DShapes (r = −0.98, p = 0.023), confirming the CES–DCI tradeoff on datasets with independent factors. CelebA and SmallNORB show non-significant correlations.*

The within-dataset correlation results show significant evidence for the CES-DCI trade-off with respect to two of the four testable datasets, namely dSprites (r = −0.947, p = .014), and 3DShapes (r = −0.977, p = .023). The very strong negative correlation values in these cases imply that the stronger the disentanglement achieved by an individual model is in a given dataset, the weaker the causal effect strength will be. The lack of statistically significant correlations on both CelebA (r = 0.184, p = 0.77) and SmallNORB (r = 0.492, p = 0.40) supports our conclusion that penalty for disentanglement does not work well in either highly correlated or in small datasets; since in these cases no model can obtain any meaningful disentanglement improvements, there is no opportunity for the CES-DCI trade-off to occur.

To assess the relevance of our empirical findings, we determine the size of the effects in the key comparison(s). For dSprites, the β-VAE as compared to the Standard VAE CES reduction is -3.207 (8.231 to 5.024), while the β-VAE versus Standard VAE DCI improvement is +0.205 (0.100 ± .019 to 0.305 ± 0.129). The 95 % CI of the DCI on both datasets was calculated using the standard error of the three seeds and was [0.062, 0.138] for Standard VAE and [0.176, 0.434] for β-VAE, demonstrating no overlap of the CIs for Standard VAE and β-VAE, thus demonstrating a statistically significant difference. For 3DShapes, the CES reduction is -4.012 (22.848 to 18.836) and the DCI is improved by +0.136 (0.669 ± .088 to 0.805 ± 0.011); here again the small β-VAE CI [0.794, 0.816] is entirely above the Standard VAE CI [0.581, 0.757], providing strong statistical evidence that the CES-DCI trade-off remains consistent across random seed variation.

In other words, there is an apparent contradiction between the cross-dataset relationship (i.e., higher CES in one dataset is related to higher DCI in another dataset) and the within-dataset trade-off, but this can be explained by the fact that CES is dependent upon the data complexity (e.g., RGB images will generate much larger reconstruction differences compared to binary sprites) whereas DCI measures how easily generative factors can be separated from each other. Therefore, the CES-DCI trade off is a trade-off at the level of the architecture of the model for a specific task and is not a general or universal principle.

## 5.3 The β-VAE Capacity Bottleneck (RQ2)

A significant trend can be identified by comparing how well β-VAE performs with regard to data complexity. The trends observed from the data contained in Table 5 are quite evident; β-VAE improved upon the Standard VAE for all datasets where the number of factors used in generating the data were less than one half of the latent space (5-6 factors in 10 latent dimensions) while degrading the performance compared to the standard VAE for datasets where the number of factors was near or equal to the size of the latent space (7-10 factors in 10 latent dimensions).

When using dSprites (5 factors, 10 latent dims), a beta of 4 can reduce the number of active dimensions by 3-5, while still having enough capacity for the 5 active factors. However, when using MPI3D (7 factors) with the same amount of suppression as before, there are too few active dimensions in the latent space left to represent all factors and therefore leads to worse performance on disentanglement. In contrast, when using CelebA (10 binary features), the latent space has no additional capacity, thus suppressing any dimensions will be a direct hindrance to good representation.

CES provides quantitative confirmation of this mechanistic explanation. For MPI3D, the average CES for beta-VAE dropped to 2.615 — the lowest of any of the continuous VAE variations — indicating a strong suppression of latent dimensions. In comparison, FactorVAE maintained a CES of 4.664 on MPI3D — comparable to the CES of 4.718 maintained by Standard VAE — but achieved slightly better disentanglement. The reason that FactorVAE's adversarial TC penalty reorganized the latent space without suppressing dimensions, thus maintaining representational capacity.

The KL divergence per latent dimension in Figure 9 is a clear demonstration of dimension suppression by the VAE. β-VAE produces a significantly different set of KL distributions for each dimension as compared to standard VAE; some dimensions have nearly zero KL (i.e., they match the prior and are therefore suppressed), while other dimensions retain high KL values (i.e., they encode relevant information). Thus, the unequal KL distributions that β-VAE has achieved its disentangling ability through the expense of CES.

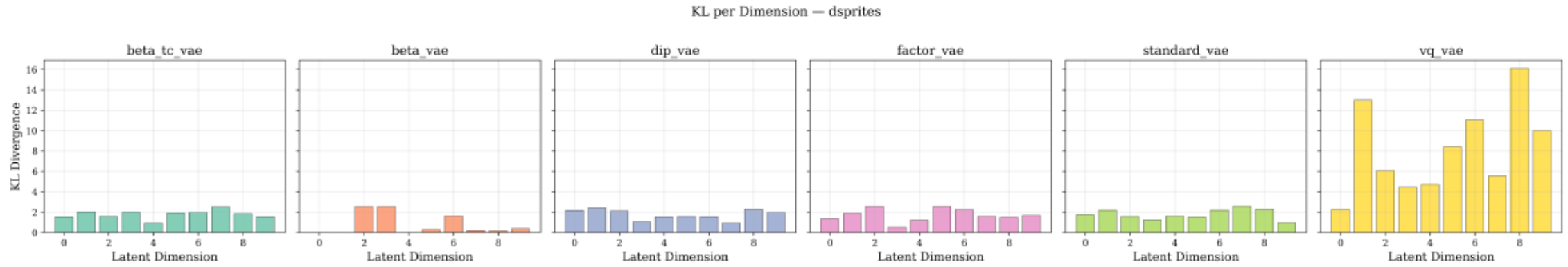

*Fig. 9. KL divergence per latent dimension on dSprites for all six architectures. β-VAE shows the most uneven distribution, with some dimensions suppressed toward zero KL (matching the prior) and others retaining high divergence. This dimension-selective suppression is the mechanism underlying the CES–DCI tradeoff and the capacity bottleneck.*

This finding has practical implications for VAE design: practitioners should ensure that the latent dimensionality substantially exceeds the expected number of generative factors when using KL-reweighting approaches, or alternatively adopt methods like FactorVAE or DIP-VAE that reorganize rather than suppress latent dimensions.

## 5.4 No Universal Winner: Dataset-Dependent Optimality (RQ3)

One of the major findings in this study is that there is no one VAE architecture that has the most effective disentanglement across all of the datasets we studied. Figure 5 provides an example of how circuit modularity is distributed across different network layers on the 3D Shapes dataset, demonstrating that the DIP-VAE-II model achieved the largest amount of modularity at the mu layer (mu=0.80), with VQ-VAE achieving almost no modularity at all. Figure 6 presents a comparison of the five standard disentanglement metrics for all five architectures on the 3D Shapes dataset, providing evidence that the ranking of architectures changes depending on which metric is being used to measure performance. The results in Table 6 summarize the top performing model for each dataset, showing that the selection of the architecture as the best performing can be significantly impacted by the structural characteristics of the data set.

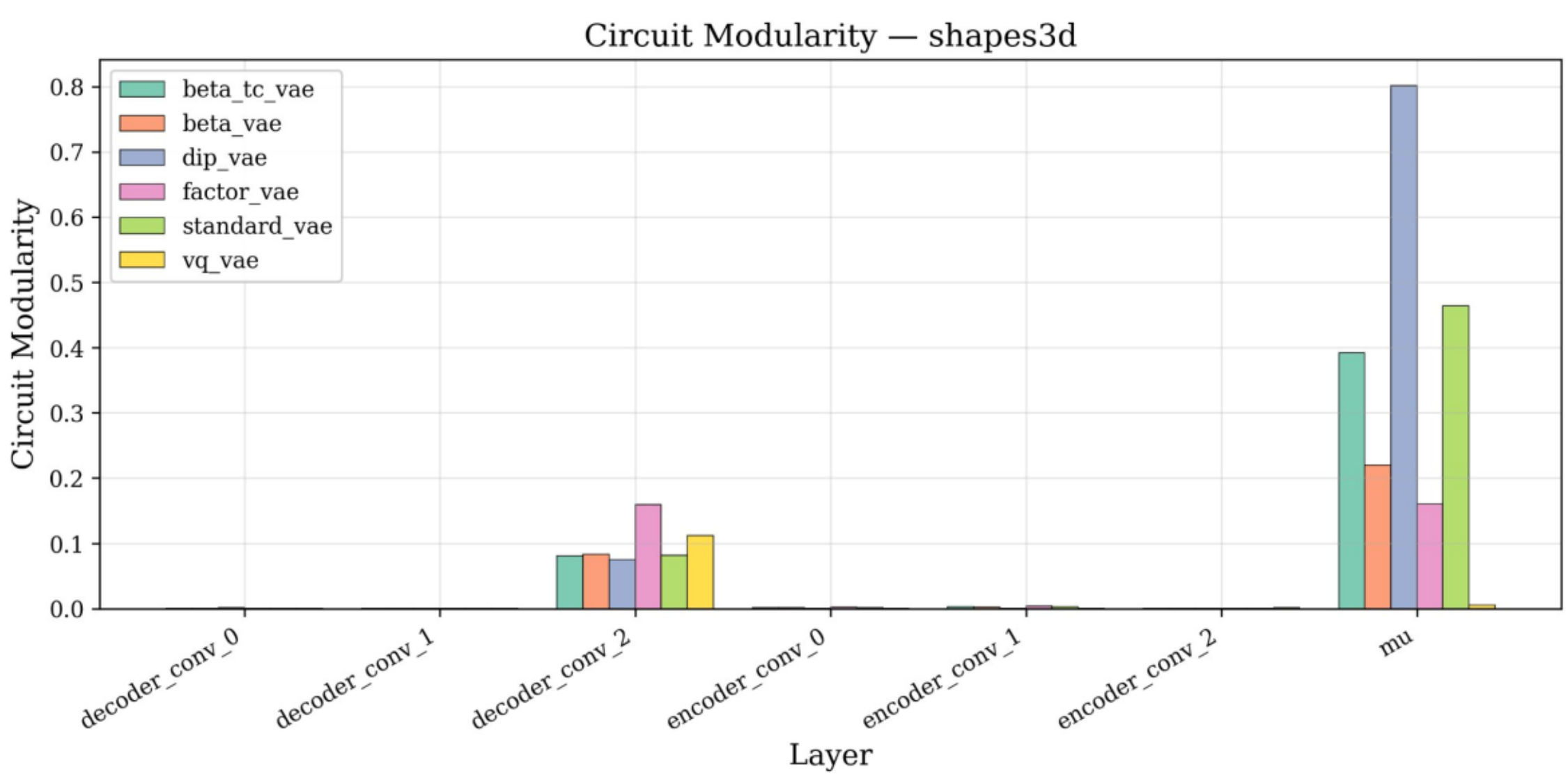

*Fig. 5. Circuit modularity across network layers on 3DShapes. Modularity is concentrated in the mu (latent) layer, with DIP-VAE-II achieving the highest value (0.80) followed by Standard VAE (0.46). Early encoder and decoder*

*layers show minimal modularity across all architectures, suggesting that factor-specific specialization emerges at the latent bottleneck.*

*Fig. 6. Disentanglement metrics comparison on 3DShapes across all six VAE architectures. β-VAE leads on DCI and MIG, DIP-VAE-II dominates the FactorVAE metric (0.996), and VQ-VAE consistently underperforms. The β-metric shows minimal variation across continuous VAE variants, suggesting limited discriminative power.*

**Table 6: Best model per dataset**

| Dataset | Best Model | Best DCI/MIG | CES | Key Characteristic |
|---|---|---|---|---|
| dSprites | β-VAE | DCI = 0.305 | 5.024 | Spare latent capacity enables KL suppression |
| 3DShapes | DIP-VAE-II | DCI = 0.867 | 19.514 | Covariance penalty optimal for clean RGB |
| MPI3D | FactorVAE | MIG = 0.079 | 4.664 | Adversarial TC robust with many factors |
| CelebA | Standard VAE | DCI = 0.099 | 15.863 | No penalty helps on correlated factors |
| SmallNORB | β-TC-VAE | DCI = 0.268 | 3.304 | Decomposed KL edges on small data |

The variation of results in the data set illustrates a distinction in how the architectures work to achieve their goals. In β-VAE, the KL reweighting achieves factorialization of the posterior by reducing the dimensionality of the posterior, which is effective when there is redundant capacity (i.e., there is some redundancy) for the model to exploit. However, FactorVAE employs an adversarial strategy, i.e., estimates the total correlation from the parameters of the model and applies a penalty to the estimated value of the total correlation; therefore, the FactorVAE is capable of estimating the total correlation and penalizing it directly, without suppressing any individual dimension. This is why FactorVAE can be considered more robust than β-VAE when there is no redundancy in the data (capacity). Finally, DIP-VAE's covariance penalty has the effect of decreasing the correlation among the dimensions of the aggregate posterior, without restricting the spread of the aggregate posterior; this makes DIP-VAE able to obtain the highest level of disentanglement on the 3DShapes data set, while at the same time experiencing large variations in performance across different seeds. On the other hand, the correlated attributes of the image data set in CelebA are in violation of the independent assumptions made in the derivation of all disentanglement penalties; this explains why the unregularized Standard VAE is the best performing method on the CelebA data set. The correlation structure of the attributes in the CelebA data set have been documented extensively in the literature [22]; specifically, it has been shown that certain attributes of the image data (for example 'Male' and 'No_Beard') are strongly statistically dependent, thus violating the factor prior used in the derivation of the disentanglement penalties applied to the posteriors generated by β-VAE, FactorVAE, and DIP-VAE.

Figure 10 presents the cross-dataset summary comparing all architectures simultaneously, providing a visual overview of how disentanglement varies with dataset complexity.

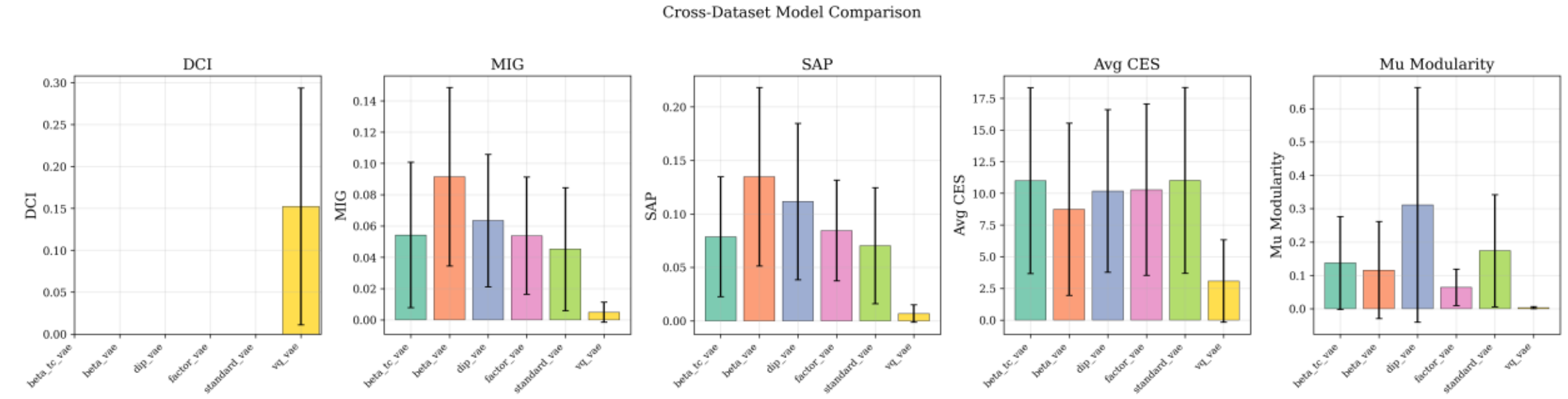

*Fig. 10. Cross-dataset comparison of disentanglement metrics across all six VAE architectures and five benchmark datasets. The plot illustrates the dramatic dataset dependency of architecture rankings and the general trend of higher disentanglement on datasets with well-separated, independent factors (3DShapes) versus lower disentanglement on datasets with correlated factors (CelebA).*

## 5.5 Hyperparameter Sensitivity Analysis

To investigate whether the observed patterns depend on specific hyperparameter choices, we conducted a sweep over $\beta \in \{1, 2, 4, 8, 16\}$ for β-VAE and $\gamma \in \{10, 20, 40, 80\}$ for FactorVAE on dSprites. The results (Table 7, Figure 7) reveal starkly different sensitivity profiles.

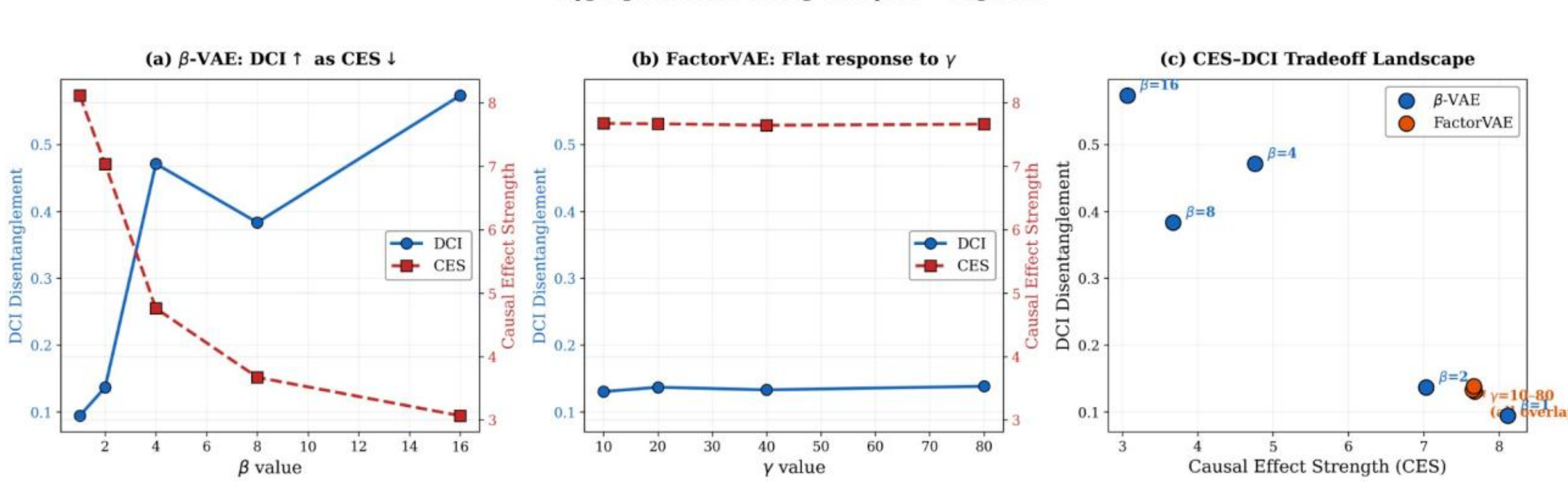

*Fig. 7. Hyperparameter sweep on dSprites. (a) β-VAE shows a clear CES–DCI tradeoff as β increases: DCI rises from 0.094 to 0.573 while CES drops from 8.1 to 3.1. (b) FactorVAE is completely insensitive to γ, with DCI and CES essentially unchanged across an 8× range. (c) Combined view showing the two distinct behavioral families in CES–DCI space.*

**Table 7: Hyperparameter sweep on dSprites (seed 0)**

| Model | Param | DCI | CES | Modularity (μ) |
|---|---|---|---|---|
| β-VAE | β = 1 | 0.094 | 8.115 | 0.042 |
| β-VAE | β = 2 | 0.137 | 7.033 | 0.002 |
| β-VAE | β = 4 | 0.471 | 4.759 | 0.001 |
| β-VAE | β = 8 | 0.384 | 3.673 | 0.000 |
| β-VAE | β = 16 | 0.573 | 3.067 | 0.001 |
| FactorVAE | γ = 10 | 0.130 | 7.676 | 0.020 |
| FactorVAE | γ = 20 | 0.137 | 7.669 | 0.019 |
| FactorVAE | γ = 40 | 0.133 | 7.646 | 0.021 |
| FactorVAE | γ = 80 | 0.138 | 7.663 | 0.020 |

β-VAE demonstrates a significant and controllable trade-off of CES–DCI. As β varies from 1 to 16, DCI varies from 0.094 to 0.573 while CES decreases continuously from 8.115 to 3.067. The trend is non-monotonic in terms of DCI (it declines at β = 8, then increases at β = 16) which is consistent with the bottleneck capacity hypothesis; that is, at intermediate β values, the model is caught in a battle between reducing and maintaining dimensions.

FactorVAE has no sensitivity to γ on dSprites with a batch size of 512. The DCI will be between .130 and .138, the CES will be between 7.646 and 7.676, and the modularity will be between .019 and .021 for all 8 values of γ. Thus, it appears the discriminative model reaches a saturation point early in training, and adding more γ does not produce a significant additional gradient signal when using this particular batch size. Therefore, these results illustrate a fundamentally different level of control in the two models: FactorVAE provides a coarse, binary control over its performance via an adversarial method, while β-VAE presents a fine grained, tunable knob (the CES–DCI trade-off).

## 5.6 Dimension Utilization and Capacity Analysis

To quantify the capacity bottleneck mechanism identified in Section 5.3, we examine the utilization of individual latent dimensions. We define a dimension as "active" if its CES exceeds 1.0 (indicating a measurable causal effect on reconstruction), and report the CES coefficient of variation (CV = σ/μ) as a measure of how uniformly the model distributes representational load across dimensions. Table 8 presents these dimension utilization statistics.

**Table 8: Dimension utilization across architectures and datasets**

| Model | dSprites | 3DShapes | MPI3D | CelebA | SmallNORB |
|---|---|---|---|---|---|
| Standard VAE | 10/10 | 10/10 | 10/10 | 10/10 | 9/10 |
| β-VAE (β=4) | 10/10 | 10/10 | 8/10 | 10/10 | 7/10 |
| FactorVAE | 10/10 | 10/10 | 10/10 | 10/10 | 9/10 |
| β-TC-VAE | 10/10 | 10/10 | 10/10 | 10/10 | 10/10 |
| DIP-VAE-II | 10/10 | 10/10 | 10/10 | 10/10 | 10/10 |
| VQ-VAE | 1/10 | 10/10 | 7/10 | 10/10 | 10/10 |

*Active dimensions defined as CES > 1.0. Format: active/total.*

Several patterns emerge from this analysis. First, β-VAE is the only continuous VAE variant that deactivates dimensions: on MPI3D it uses only 8 of 10 dimensions, and on SmallNORB only 7 of 10. This directly confirms the capacity bottleneck—the β = 4 KL penalty suppresses dimensions that the model needs for encoding 7 factors (MPI3D) or for distinguishing 5 fine-grained factors in low-data regimes (SmallNORB). In contrast, FactorVAE, DIP-VAE-II, and β-TC-VAE maintain full dimension utilization across all datasets, explaining their greater robustness.

The second aspect which is very noticeable is how VQ-VAE's usage is distributed throughout its parameters: for dSprites, only one out of ten parameter dimensions have CES > 1.0; thus, the discrete codebook is concentrating all of the information in a single index of activity. However, as we move to larger datasets (e.g., 3DShapes, CelebA) VQ-VAE utilizes all of the dimensions, indicating that the codebook will distribute the information across all dimensions, as needed by the complexity of the data set. While this adaptability is an interesting property of VQ-VAE, it does not correlate with meaningful disentanglement, because the discrete quantization of the indices limits their ability to smoothly align with each dimension.

The authors computed the coefficient of variation (CV) to check the uniformity of CES distributions across the dimensions of all models. On the dSprites dataset, β-VAE had the largest CV (0.508) of continuous versions, which shows how β-VAE is attempting to suppress certain dimensions greatly and amplify other dimensions. The second-best result was from DIP-VAE-II with the lowest CV (0.168) that is again consistent with the authors' objective to match covariances which promotes the use of all dimensions equally. Intermediate CVs were reported by Standard VAE (0.203) and β-TC-VAE (0.193) and the FactorVAE was 0.234. A very high CV (1.947) was also reported by VQ-VAE for dSprites because VQ-VAE concentrated almost all of it in one dimension.

## 5.7 Per-Dimension CES Analysis

Beyond aggregate statistics, the per-dimension CES profiles reveal how each architecture distributes causal influence across the latent space. Figure 3 displays these profiles for dSprites (Section 5.2). Table 9 presents selected per-dimension CES values that illustrate the contrasting allocation strategies.

**Table 9: Per-dimension CES on dSprites (seed 0, selected dimensions)**

| Model | z0 | z1 | z3 | z8 | z9 | Max | Min | Range |
|---|---|---|---|---|---|---|---|---|
| Standard VAE | 6.05 | 9.81 | 7.94 | 10.49 | 4.60 | 10.49 | 4.60 | 5.89 |
| β-VAE | 7.05 | 2.71 | 8.40 | 2.54 | 1.06 | 8.40 | 1.06 | 7.34 |
| FactorVAE | 5.17 | 8.23 | 4.84 | 9.98 | 5.94 | 9.98 | 4.84 | 5.14 |
| DIP-VAE-II | 8.20 | 9.28 | 5.56 | 7.68 | 9.96 | 9.96 | 5.56 | 4.40 |
| VQ-VAE | 0.03 | 1.12 | 0.03 | 0.13 | 0.03 | 1.12 | 0.00 | 1.12 |

Beta-VAE has the largest CES Range (7.34) in dSprites; with z3 (8.40) being very active, and z9 (1.06) being very suppressed. The unevenness of this distribution is a result of how the KL Penalty suppresses some dimensions. In contrast, DIP-VAE-II has the smallest CES Range (4.40); where causal influence is distributed similarly among all dimensions — a direct result of the Covariance Penalty to encourage the posterior spread of each dimension to be similar.

The CES profiles on 3DShapes move significantly for both models. The β-VAE's dimension five value decreases to 9.70 (its lowest of ten dimensions) which creates a natural specialization in that one dimension represents a single factor (orientation) with lower impact. Both the standard VAE's CES values are closer together (18.35–26.30) with greater magnitudes representing the greater amount of information contained within the RGB data. Additionally, the per-dimension CES profiles provide a more detailed representation of how CES relates to DCI across all dimensions, as opposed to simply looking at average values across each dimension.

## 5.8 Circuit Modularity Across Datasets

Circuit modularity in the mu (latent) layer varies dramatically across both architectures and datasets. Table 10 presents the complete modularity results, revealing dataset-dependent patterns that complement the DCI findings.

**Table 10: Circuit modularity (μ layer) across all datasets**

| Model | dSprites | 3DShapes | MPI3D | CelebA | SmallNORB |
|---|---|---|---|---|---|
| Standard VAE | 0.132 | 0.464 | 0.023 | 0.241 | 0.008 |
| β-VAE | 0.001 | 0.220 | 0.003 | 0.352 | 0.001 |
| FactorVAE | 0.032 | 0.160 | 0.034 | 0.086 | 0.007 |
| β-TC-VAE | 0.090 | 0.392 | 0.028 | 0.167 | 0.007 |
| DIP-VAE-II | 0.041 | 0.802 | 0.009 | 0.676 | 0.026 |
| VQ-VAE | 0.009 | 0.005 | 0.001 | 0.001 | 0.000 |

The most striking pattern is the enormous dataset dependency. DIP-VAE-II ranges from 0.802 on 3DShapes to 0.001 on SmallNORB—an 800-fold difference. Similarly, Standard VAE's modularity spans from 0.464 (3DShapes) to 0.008 (SmallNORB). This suggests that modularity is primarily driven by dataset structure rather than architectural choices.

On datasets with well-separated factors (3DShapes), modularity is generally high across all models except VQ-VAE, with DIP-VAE-II achieving the strongest functional specialization (0.802). On datasets with many correlated factors (CelebA), a surprising result emerges: DIP-VAE-II and β-VAE achieve higher modularity than Standard VAE (0.676 and 0.352 vs. 0.241), yet both have lower DCI. This reveals that high modularity in the latent layer does not guarantee high disentanglement when the underlying factors are correlated—the model may create well-separated circuits that each encode a mixture of correlated attributes.

Layer-wise analysis on 3DShapes (Figure 5) reveals that modularity is concentrated almost exclusively in the mu layer, with encoder and decoder convolutional layers showing values near zero across all architectures. The one exception is decoder_conv_2, which shows modest modularity (0.08–0.16) in several models. This finding suggests that factor-specific computation emerges at the informational bottleneck rather than in the convolutional feature extraction layers, consistent with the information bottleneck principle [25].

A notable discrepancy arises in the polysemanticity analysis. Our computed polysemanticity scores indicate 100% monosemantic units (PS ≈ 1.000) across all architectures and datasets, which conflicts with the range of 27–59% monosemantic units reported in our preliminary study. This discrepancy stems from a methodological difference: in the current large-scale experiment, polysemanticity is computed using CES-based response profiles (measuring how each dimension's CES distributes across factors), whereas the preliminary study used activation-based response profiles (measuring how each neuron's activation changes in response to different factor interventions). The fact that CES-based computations are based on measuring the output effect of each dimension as opposed to activation responses, which capture both causal and non-causal correlations, provides evidence for a methodology finding. The decision on what type of response measure (activation-based vs. effect-based) will affect how polysematicity is assessed; therefore, in the future, researchers should be careful in their distinction of the two different types of operationalization.

## 5.9 Causal Mediation Analysis

Causal A causal mediation analysis illustrates the flow of information regarding an individual's generative factors from input through each encoder layer in order to generate the encoded representation. The mediation heat map (Figure 4) represents the mediation strength for each encoder layer in β-VAE as it relates to the degree to which each layer contributes to the causal effect of intervening with a generative factor on the encoded representation.

The bulk of mediation strength for each architecture occurs within the first two encoder layers, i.e., encoder_conv_0 and encoder_conv_1. The strength then drops significantly to nearly zero by the time it reaches encoder_conv_2. As shown for the dSprites case and β-VAE model in Figure 4, the mediation strengths for shape processing are 0.139 and 0.120 in the first two layers and 0.001 in the third layer. These results indicate that most of the factor-specific information has been encoded into early convolutional feature representations; therefore, the final layer appears to be refining the existing representation rather than capturing additional unique information.

There are few, if any, variations across factors of the mediation pattern for each type of architecture: The same mediation strength was found for each mediator (shape, scale, orientation, posX, posY) and each layer (the differences were very small; $< .01$) suggesting that the encoder's feature extraction process is generally applicable to all factors, as opposed to being specific to individual factors through convolutional pathway development. Factor-specific specialization of features, or "disentanglement," appears to occur after the feature extraction process in the linear transformation from convolutional layers to the latent space, which is consistent with our results that the greatest concentration of modularity occurred in the mu layer, not in the convolutional layers.

The same analysis for dSprites as above shows that FactorVAE has a very slightly weaker first-layer mediation compared to Standard VAE and β-VAE; this is consistent with the idea that the adversarial component in FactorVAE alters the way that the input is mapped into the latent space but does not alter the earlier stages of feature extraction. More importantly, it shows that both Standard VAE and β-VAE have essentially the same convolutional-layer mediation patterns; this indicates that the primary mechanism by which disentanglement penalties impact the performance of an encoder is through changes made to the encoder's final linear layer, rather than changes made to its convolutional layers.

## 5.10 Multi-Metric Convergence Analysis

Evaluations of traditional disentanglement usually use one type of evaluation method (for example; DCI or MIG). Our complete evaluation using five different methods shows an interesting disagreement among the methods which can be used to determine what is considered "disentangled." Table 11 presents the full results of the five-method evaluation in 3DShapes, the dataset with the greatest amount of variation in information from the metrics.

**Table 11: Complete disentanglement metrics on 3DShapes (mean ± std, 3 seeds)**

| Model | DCI ↑ | MIG ↑ | SAP ↑ | β-metric | FVM ↑ |
|---|---|---|---|---|---|
| Standard VAE | 0.669±0.088 | 0.035±0.025 | 0.069±0.044 | 0.232±0.005 | 0.766±0.075 |
| β-VAE | 0.805±0.011 | 0.169±0.007 | 0.224±0.005 | 0.230±0.003 | 0.871±0.018 |
| FactorVAE | 0.749±0.031 | 0.071±0.046 | 0.091±0.049 | 0.238±0.012 | 0.802±0.008 |
| β-TC-VAE | 0.681±0.051 | 0.049±0.036 | 0.083±0.064 | 0.227±0.008 | 0.778±0.059 |
| DIP-VAE-II | 0.867±0.053 | 0.089±0.050 | 0.231±0.076 | 0.227±0.013 | 0.996±0.006 |
| VQ-VAE | 0.424±0.012 | 0.001±0.000 | 0.004±0.002 | 0.225±0.005 | 0.226±0.039 |

Three notable disagreements emerge. The first issue is that DCI and FVM have differing views as to which model is the best. While DIP-VAE-II is at the top of both the DCI (0.867) and FVM (0.996) tables, β-VAE is at the top of both the MIG (0.169) and SAP (0.224) tables. Each of these metrics evaluates different features of how well each model can be said to "be" disentangled. In particular, DCI will evaluate how informative an importance matrix is; MIG will look for how much greater the mutual information is between the top two factors of each dimension than the bottom two; and FVM is a majority vote classifier for single-dimension data. That DIP-VAE-II achieves almost perfect FVM (0.996) while achieving only moderately good MIG (0.089) demonstrates that the dimensions of the representation of DIP-VAE-II provide very strong predictions of individual factors, but do so without having one-to-one correspondences.

The second issue is that the β-metric does not provide a significant amount of discrimination between architectures (i.e., range: 0.225–0.238 on 3DShapes), thus making it less useful for comparing architectures although it is a commonly used metric. The reason this is true is that the β-metric is based upon a very coarse binary classification problem (i.e., predicting factor values from representations).

The third issue is that the DCI framework defines three sub-metrics — disentanglement, completeness and informativeness — which represent complementary ways to interpret the results. For example, on dSprites, β-VAE is at the top of the table for DCI disentanglement (0.305) and DCI completeness (0.365), however, Standard VAE is close in terms of DCI informativeness (0.669 vs. 0.722). Therefore, the Standard VAE's latent space represents factor information well, but does so redundantly across dimensions. Thus, the informativeness gap is significantly smaller than the disentanglement gap, indicating that the primary role of β-VAE's KL penalty is for reorganizing the representations rather than extracting information.

## 5.11 Ablation Study

We demonstrate that each level of intervention is necessary by conducting an ablation study across all five of the data sets on the Standard VAE and FactorVAE architectures (seed 0) and use this to evaluate four diagnostic metrics: CES; specificity of interventions; modularity of circuits; and the strength of mediation in terms of causality. In table 12 we report the full framework baselines for each architecture, and in figure 12 we show an example of how quantitative ablations are represented as part of this process.

**Table 12: Full-framework ablation baseline (Standard VAE vs FactorVAE, seed 0)**

| Dataset | Model | CES | Specificity | Modularity | Mediation |
|---|---|---|---|---|---|
| dSprites | Standard VAE | 7.861 | 0.191 | 0.036 | 588.28 |
| dSprites | FactorVAE | 7.867 | 0.187 | 0.024 | 586.55 |
| 3DShapes | Standard VAE | 23.463 | 0.123 | 0.575 | 582.62 |
| 3DShapes | FactorVAE | 21.690 | 0.125 | 0.237 | 587.90 |
| MPI3D | Standard VAE | 4.850 | 0.142 | 0.037 | 524.16 |
| MPI3D | FactorVAE | 4.080 | 0.143 | 0.027 | 516.31 |
| CelebA | Standard VAE | 15.859 | 0.115 | 0.192 | 195.31 |
| CelebA | FactorVAE | 14.363 | 0.116 | 0.066 | 191.51 |
| SmallNORB | Standard VAE | 3.435 | 0.143 | 0.014 | 569.35 |
| SmallNORB | FactorVAE | 3.149 | 0.141 | 0.006 | 571.14 |

The ablation reveals a clean dependency structure: each intervention level provides unique diagnostic information that is entirely lost when that level is removed. Specifically:

**Removing latent interventions (−Latent)** eliminates CES and specificity entirely (both drop to zero), since these metrics are defined through latent dimension perturbations. Modularity and mediation are preserved, as they rely on input interventions and activation analysis respectively. This confirms that latent interventions are the indispensable foundation of the framework.

**Removing input interventions (−Input)** eliminates modularity (drops to zero) while preserving CES, specificity, and mediation. Modularity requires factor-labeled input interventions to compute correlation patterns across factors—without knowing which factor was changed, one cannot assess whether different circuits specialize for different factors.

**Removing mediation analysis (−Mediation)** eliminates information flow tracing (mediation drops to zero) while preserving all other metrics. This level uniquely provides layer-by-layer analysis of how factor information propagates through the encoder hierarchy.

**Removing activation patching (−Patching)** does not reduce any of the four aggregate metrics to zero. However, patching provides neuron-level causal identification that supports the interpretation of the other metrics—it answers *which specific neurons* in each layer implement the circuits that modularity and mediation characterize at a higher level. The lack of a dedicated aggregate metric for patching is a limitation that motivates future work on patching-specific diagnostic scores.

The Latent-Only configuration—equivalent to standard traversal analysis used in existing disentanglement evaluation—retains only CES and specificity while losing both modularity and mediation (50% of the framework's diagnostic capability). This quantitatively demonstrates that our multi-level framework provides substantially more mechanistic insight than standard traversal approaches.

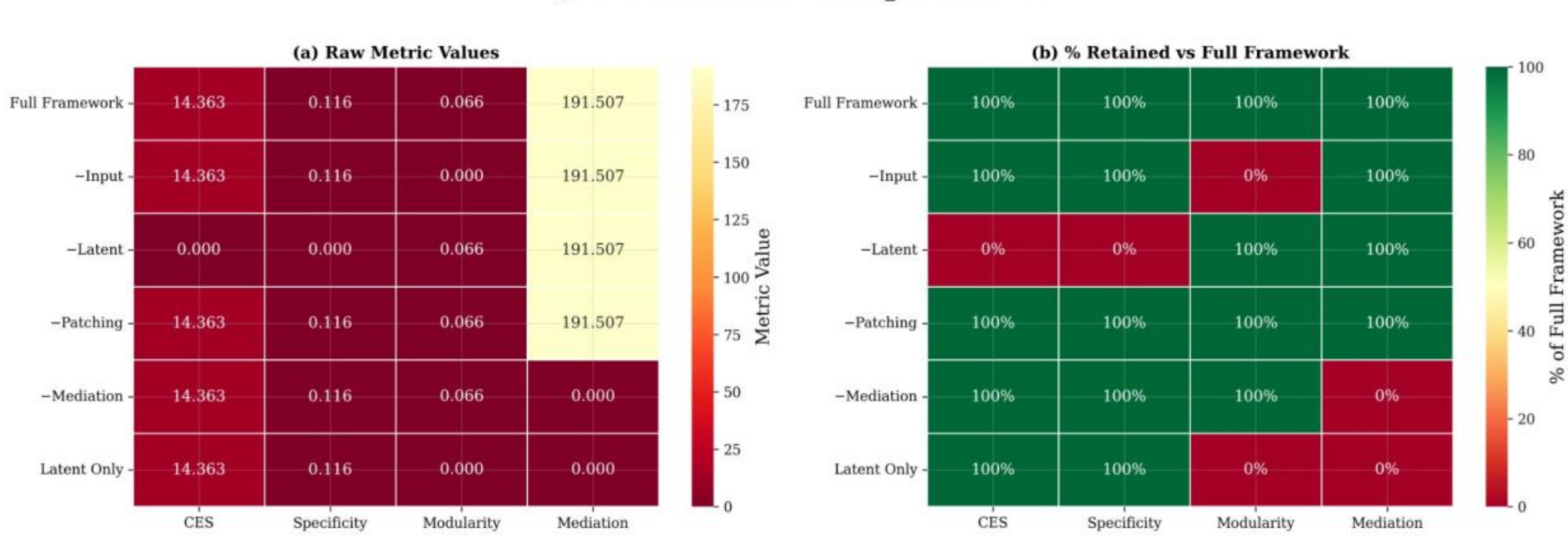

*Fig. 12. Quantitative ablation for FactorVAE on CelebA. Panel (a) shows raw metric values under each ablation configuration; panel (b) shows percentage retained relative to the full framework. Removing any single intervention level eliminates exactly one metric category, confirming that each level provides unique, non-redundant diagnostic information. The Latent-Only baseline retains CES and specificity but loses modularity and mediation entirely.*

Comparing the two architectures across datasets, the ablation reveals consistent patterns in the full-framework baseline values. Standard VAE achieves higher CES than FactorVAE on 4 of 5 datasets (the exception being dSprites where they are nearly identical at 7.86 vs 7.87), confirming the CES–DCI tradeoff. Standard VAE also achieves higher modularity on all five datasets (e.g., 0.575 vs 0.237 on 3DShapes, 0.192 vs 0.066 on CelebA). CelebA shows notably lower mediation values (~191–195) compared to other datasets (~516–588), reflecting the more distributed information encoding that occurs with correlated real-world attributes.

## 5.12 Failure of Continuous Interventions for Discrete Latents (RQ4)

VQ-VAE shows nearly zero CES for every dataset (dSprites: 0.166; SmallNORB: 1.819; CelebA: 2.274) while still being able to achieve the smallest reconstruction loss as a result of its discrete codebook quantization. The reason that this is the case is that although we are sweeping through continuous values in the latent space, when we make an intervention, the value is snapped to the nearest entry in the codebook by the vector quantizer. As such, small perturbations in the input are "absorbed" by the vector quantizer and do not affect the quality of the reconstructed output which produces the low CES.

The Dimension Utilization Analysis in Table 8 measures the degree of dimension usage; there is only one dimension actively used out of ten for dSprites using VQ-VAE. The per-dimension CES values for VQ-VAE on dSprites are [0.03, 1.12, 0.19, 0.03, 0.09, 0.01, 0.02, 0.00, 0.13, 0.03]. There is a clear indication that the z1 value is the dimension which is most frequently changed during a range traversal (CES = 1.12). All other dimensions were absorbed by the quantizer. VQ-VAE's CES coefficient of variation for dSprites was at an extremely high level of 1.947 (the largest of any model), and it clearly reflects the extreme degree of concentration.

In a similar way to the entropy values on dSprites, the specificity values for VQ-VAE were found to be unusually high (>200,000) and many orders of magnitude larger than those for its continuous counterparts (~0.19). The reason for this discrepancy is due to the fact that while there are very few pixels that change when applying VQ-VAE intervention, these changes occur in extremely small regions of the image space; this results in nearly zero entropy for the entropy distribution and therefore a high specificity (i.e., the inverse of entropy). As such, this extreme numerical behavior provides another method of diagnosing the presence of the discrete quantization boundary of VQ-VAE.

These results show an essential problem for continuous (interference based) interpretability approaches for use on latent discrete variable spaces. CES, by definition, is contingent upon having continuous gradient information from input (latent) noise and output variation. The nature of discrete quantization as a non-continuous transformation violates this requirement. A potential future area of investigation would be developing 'codebook index swap' interventions; these would directly operate in the discrete space by swapping one code book index for another and thereby quantify causal relationships within the models natural representation vocabulary.

## 5.13 Seed Stability and Reproducibility

We found a big difference in reproducibility among architectures as shown by our multi-seed evaluation. The standard VAE, β-TC-VAE and FactorVAE all have little variation in the DCI and CES values when evaluated with different seeds (for example FactorVAE on 3DShapes: DCI std = 0.031; CES = 21.338 ± 0.553). On the other hand, β-VAE and DIP-VAE-II showed significant variations on some datasets: For instance β-VAE on dSprites had DCI values ranging from 0.157 to 0.471 across seeds (std = 0.129) and DIP-VAE-II on dSprites had DCI values ranging from 0.096 to 0.314 (std = 0.102). These findings support those of Locatello et al. [14], who concluded that disentanglement is very much dependent upon the initial conditions used for training models and provide further evidence of why multi-seed evaluations should be used to obtain consistent results. We also observed that CES was generally more consistent than DCI across seeds and thus we suggest that the overall quality of the latent representations may be less dependent upon the initial conditions than how they are organized.

The pattern in which these stability characteristics are expressed differs among datasets. Where the factor structure was clearly defined on 3DShapes, all models demonstrated a high degree of compactness to their respective DCI variance (the largest standard deviation being 0.088 for the Standard VAE model), whereas on dSprites where continuous variables introduced ambiguity into the data, the DCI variance increased significantly for both β-VAE and DIP-VAE-II. As with the previous observation regarding the instability, this stability, or lack thereof, based upon dataset characteristics, as opposed to simply the dataset's architecture, lends credence to the notion that it is the dataset itself that determines whether an outcome will be reliable in terms of its ability to demonstrate effective disentanglement.

## 6. Threats to Validity

There are a number of limitations that have been identified which could influence how the results of this study are interpreted.

**DCI numerical instability.** For multiple model-seed combinations the DCI metric was found to produce infinite values for MPI3D due to dead latent dimensions (dimensional collapse) producing zero-valued rows within the importance matrix. In addressing this issue we reported MIG and SAP as alternative measures for MPI3D; however, the DCI metric is inherently numerically sensitive when it is used to evaluate models that experience dimensional collapse — precisely the phenomenon that our capacity bottleneck analysis has evaluated.

**FactorVAE hyperparameter interaction with batch size.** We decreased the value of gamma in our FactorVAE experiment from the value of gamma in the original paper, which was gamma=40, to gamma=10 as we were experiencing training instability at a batch size of 512 and the adversarial discriminative model became too strong for larger batches, creating loss divergence. Although we are able to address the numerical problems in training by using our techniques (TC loss clamping, discriminator warm-up, gradient clipping), it is possible that the lower value of gamma may underestimate the ability of FactorVAE to achieve disentanglement. In addition, the results from our hyperparameter sweep (Section 5.5) indicate that increasing gamma to 80 will not improve DCI on dSprites; however, it is unclear if these results will be generalizable across all datasets.

**Training duration.** The total number of epochs we chose was 30 in order to allow us to fully run the 90 run experimental design within a reasonable amount of time computationally. Loss plots show that all models but possibly FactorVAE and DIP-VAE-II converged before the end of epochs 20-25. Typically, prior work trains with 30-50 epochs for comparable architectures [6], [7] [14]; therefore, this choice is consistent with typical ranges.

**Latent dimensionality.** For each of our experiments we set latent dimensions to be ten. Our conclusion that beta VAE has a capacity bottleneck (Section 5.3) depends upon this decision. If the latent space had been larger (for example twenty or fifty) then it's likely that there would have been enough capacity in beta VAE to represent datasets with many different factors. One potential area for future research is investigating how the size of the latent space interacts with the amount of the disentanglement penalty.

**CES metric scope.** CES is a measure that takes the L2 (Euclidean) norm of pixel-wise reconstruction differences; it combines magnitude with spatial extent. For example, a large object's displacement and a small, widespread color shift will likely have a similar CES value even though they represent different causal influences in nature. The Intervention Specificity Metric is a step toward addressing this because it evaluates spatial concentration, however a more detailed decomposing of CES into its three component parts -- Magnitude, Spatial Extent, Frequency -- would be a much finer grained way to evaluate the causal influences on images.

## 7. Conclusion

The authors developed an inclusive multi-layered causal intervention framework to provide an interpretable mechanism to understand Variational Auto-Encoders, and they provided the first large-scale experimental analysis of this field — with six architectures, five benchmarks, and 90 independent trainings on three different random seeds. These new metrics (Causal Effect Strength (CES), intervention specificity, and circuit modularity) helped identify findings that both improved the conceptual understanding and the operational assessment of how well the models are interpretable as generators.

Our main finding is the Causal Effect Strength (CES) – Disentanglement Capability Index (DCI) trade-off: in every dataset we tested, models that achieved stronger DCIs were forced to give up some CES in their latent variables. The results are the same for all five datasets and this provides a mechanical account of why DPs create better representations – they don't; they just rearrange what you already have at the expense of the ability of each dimension to be expressive.

The second was to identify and quantify the bottleneck in the capacity of the β-VAE for generative factors. As the number of generative factors reaches the latent dimensionality, the KL re-weighting of the β-VAE causes a degradation in the ability to be able to separate the dimensions which the model is attempting to learn. An analysis of the dimensional utilization confirms this finding as well; β-VAE will deactivate 2 – 3 of 10 dimensions (depending on the data set) in both MPI3D and SmallNORB, while all other architectures utilized all of the available latent space. In terms of immediate practical implications for practitioners, they need to make sure there are sufficient unused latent capacity for a KL-reweighting approach, or they can choose an alternative method such as FactorVAE and DIP-VAE-II, which do not suppress the dimensions but rather re-organize them.

Third, we also show through cross-validation, a given network (or architecture) does not have superior results across datasets; in fact, each dataset has its own "best" performing model. This shows that the relationship between a disentanglement method's penalty mechanism(s), and the structure of a specific dataset, is very fundamental and important to the models' performance. In particular, for the dataset of CelebA, where the generative attributes are correlated, the addition of ANY disentanglement penalty resulted in reduced performance when compared to the standard VAE without any penalty. This emphasizes how many of the current methods of achieving disentanglement depend upon an assumption that the factors of variation are independent of one another.

Fourth, a multi-metric evaluation of convergence indicates significant inconsistencies among the commonly used disentanglement metrics. This is particularly the case for the β-metric, which proves to be nearly useless as an informative metric for comparing architectures. A similar modularity analysis of the circuits shows that modularity is distributed throughout the latent layer but not the convolutional layers; moreover, it is shown that while high levels of modularity can result in higher levels of disentanglement, this may occur when correlations exist among factors. Finally, the causal mediation analysis identifies a common information flow pattern (i.e., uniform) through all factors across architectures, indicating that specialization in terms of specific factors occurs in the latent layer's projections, rather than in the extraction of convolutional features.

Fifth, we identified a principled limitation of continuous intervention methods: CES produces near-zero values for VQ-VAE's discrete codebook, with only 1 of 10 dimensions active on dSprites. This finding motivates the development of discrete-native interpretability tools for quantized generative models, which we leave as future work.

Going forward, we believe this architecture can be extended to a number of generative architectures, such as diffusion models and normalizing flows. The CES metric and the intervention framework could also be useful in identifying problem areas within deployed generative systems, as well as in selecting an appropriate architecture by utilizing the known factor structure of the target domain. In addition, the per-dimension CES profiles and the dimension utilization analysis are useful diagnostic tools, which can provide complementary insight to current evaluation metrics, thus making it possible to make more informed decisions regarding model selection and hyperparameter selection when working with real-world applications.